%% file: main.tex
\documentclass[letterpaper,10pt,journal,twoside]{IEEEtran}
\input{macros.tex}

\newcommand\copyrighttext{%
  \scriptsize \textcopyright 2024 IEEE. Personal use of this material is permitted.
  Permission from IEEE must be obtained for all other uses, in any current or future
  media, including reprinting/republishing this material for advertising or promotional
  purposes, creating new collective works, for resale or redistribution to servers or
  lists, or reuse of any copyrighted component of this work in other works.
  DOI: \href{https://doi.org/10.1109/LRA.2024.3364842}{10.1109/LRA.2024.3364842}}
\newcommand\copyrightnotice{%
\setlength{\fboxsep}{2pt}
\begin{tikzpicture}[remember picture,overlay]
\node[anchor=south,yshift=10pt] at (current page.south) {\fbox{\parbox{\dimexpr\textwidth-\fboxsep-\fboxrule\relax}{\copyrighttext}}};
\end{tikzpicture}%
}

\title{CLIPPER: Robust Data Association \\%
without an Initial Guess}

\author{Parker C. Lusk and Jonathan P. How%
    \thanks{Manuscript received: September, 29, 2023; Revised December, 26, 2023; Accepted January, 23, 2024.}%
    \thanks{This paper was recommended for publication by Editor Sven Behnke upon evaluation of the Associate Editor and Reviewers' comments.
This work was supported by the Ford Motor Company and ARL DCIST under Cooperative Agreement W911NF-17-2-0181, and by UPenn under ONR award 584551.} %
	\thanks{P.\ C.\ Lusk, J.\ P.\ How are with the Department of Aeronautics and Astronautics, Massachusetts Institute of Technology.
	    {\{plusk, jhow\}@mit.edu.}}
}%

\newcommand{\rev}[1]{#1}

\begin{document}

\maketitle
\copyrightnotice

\markboth{IEEE Robotics and Automation Letters. Preprint Version. Accepted January, 2024}
{Lusk and How: CLIPPER: Robust Data Association without an Initial Guess}

\begin{abstract} 
\input{paper/abstract}
\end{abstract}

\begin{IEEEkeywords}
Optimization and Optimal Control; Sensor Fusion; RGB-D Perception; Localization; Mapping
\end{IEEEkeywords}

\input{paper/intro}

\input{paper/related_work}

\input{paper/consistent_measurement_selection}

\input{paper/clipper}
\input{paper/experiments}
\input{paper/conclusion}

\balance %

\bibliographystyle{IEEEtranN} %
\bibliography{refs}

\end{document}

%% file: macros.tex
\usepackage{amsmath} %
\usepackage{amssymb} %
\usepackage{amsfonts}
\usepackage[usenames,dvipsnames]{xcolor}
\usepackage{graphicx}
\usepackage{tabularx}
\usepackage{multirow}
\usepackage{mathtools}
\usepackage{esvect} %
\usepackage[binary-units=true]{siunitx}
\usepackage{caption}
\usepackage[pdftex, pdfstartview={FitV}, pdfpagelayout={TwoColumnLeft},bookmarksopen=true,plainpages = false, colorlinks=true, linkcolor=black, citecolor = black, urlcolor = black,filecolor=black , pagebackref=false,hypertexnames=false, plainpages=false, pdfpagelabels ]{hyperref}
\usepackage[T1]{fontenc} %
\usepackage{mathtools, cuted} %
\usepackage{bm}
\usepackage{xspace}
\usepackage{stackengine}

\usepackage{balance} %
\usepackage{booktabs} %
\usepackage[export]{adjustbox} %

\usepackage[numbers]{natbib} %

\newcolumntype{x}[1]{%
>{\raggedleft\hspace{0pt}}p{#1}}%

\usepackage{algorithm}
\usepackage[noend]{algpseudocode} %
\definecolor{commentclr}{RGB}{34, 139, 34} %
\newcommand{\algcomment}[1]{{\color{commentclr}// #1}}

\usepackage{tikz}
\usetikzlibrary{plotmarks}
\usetikzlibrary{positioning}
\usetikzlibrary{shapes}
\newcommand{\tikzhexagon}[1][0.1]{\protect\tikz{\node[regular polygon, regular polygon sides=6, fill=red, scale=0.6] {}; }}%

\usepackage{subcaption}
\DeclareCaptionLabelSeparator{periodspace}{.\quad}
\newcommand{\ra}[1]{\renewcommand{\arraystretch}{#1}}

\usepackage{amsthm}

\theoremstyle{definition}

\usepackage{letltxmacro}
\LetLtxMacro\orgvdots\vdots
\LetLtxMacro\orgddots\ddots

\makeatletter
\DeclareRobustCommand\vdots{%
	\mathpalette\@vdots{}%
}
\newcommand*{\@vdots}[2]{%
	\sbox0{$#1\cdotp\cdotp\cdotp\m@th$}%
	\sbox2{$#1.\m@th$}%
	\vbox{%
		\dimen@=\wd0 %
		\advance\dimen@ -3\ht2 %
		\kern.5\dimen@
		\dimen@=\wd2 %
		\advance\dimen@ -\ht2 %
		\dimen2=\wd0 %
		\advance\dimen2 -\dimen@
		\vbox to \dimen2{%
			\offinterlineskip
			\copy2 \vfill\copy2 \vfill\copy2 %
		}%
	}%
}
\DeclareRobustCommand\ddots{%
	\mathinner{%
		\mathpalette\@ddots{}%
		\mkern\thinmuskip
	}%
}
\newcommand*{\@ddots}[2]{%
	\sbox0{$#1\cdotp\cdotp\cdotp\m@th$}%
	\sbox2{$#1.\m@th$}%
	\vbox{%
		\dimen@=\wd0 %
		\advance\dimen@ -3\ht2 %
		\kern.5\dimen@
		\dimen@=\wd2 %
		\advance\dimen@ -\ht2 %
		\dimen2=\wd0 %
		\advance\dimen2 -\dimen@
		\vbox to \dimen2{%
			\offinterlineskip
			\hbox{$#1\mathpunct{.}\m@th$}%
			\vfill
			\hbox{$#1\mathpunct{\kern\wd2}\mathpunct{.}\m@th$}%
			\vfill
			\hbox{$#1\mathpunct{\kern\wd2}\mathpunct{\kern\wd2}\mathpunct{.}\m@th$}%
		}%
	}%
}
\makeatother

\let\oldnl\nl%
\newcommand{\nonl}{\renewcommand{\nl}{\let\nl\oldnl}}%
\makeatother

\newcommand{\Real}[1]{ { {\mathbb R}^{#1} } }

\newcommand{\SO}[1]{\ensuremath{\mathrm{SO}(#1)}\xspace}

\newcommand\eqdef{\mathrel{:=}}

\graphicspath{{figures/}}

\usepackage{svg}
\svgpath{{figures/}}

%% file: paper/abstract.tex
Identifying correspondences in noisy data is a critically important step in estimation processes.
When an informative initial estimation guess is available, the data association challenge is less acute; however, the existence of a high-quality initial guess is rare in most contexts.
We explore graph-theoretic formulations for data association, which do not require an initial estimation guess.
Existing graph-theoretic approaches optimize over unweighted graphs, discarding important \emph{consistency} information encoded in weighted edges, and frequently attempt to solve NP-hard problems exactly.
In contrast, we formulate a new optimization problem that fully leverages weighted graphs and seeks the densest edge-weighted clique.
We introduce two relaxations to this problem: a convex semidefinite relaxation which we find to be empirically tight, and a fast first-order algorithm called CLIPPER which frequently arrives at nearly-optimal solutions in milliseconds.
When evaluated on point cloud registration problems, our algorithms remain robust up to at least 95\% outliers while existing algorithms begin breaking down at 80\% outliers.
Code is available at \href{https://mit-acl.github.io/clipper}{https://mit-acl.github.io/clipper}.

%% file: paper/intro.tex
\section{Introduction}\label{sec:intro}

\IEEEPARstart{D}{ata} association is a fundamental requirement of geometric estimation in robotics.
Identifying correspondences between measurements and models enables estimation processes to incorporate more data, in general leading to better estimates.
However, sensor data is replete with noise and spurious measurements, making data association considerably more challenging.
Further, while some estimation processes (e.g., ICP~\cite{besl1992method}) leverage an initial estimation guess to reduce the difficulty of data association, high-quality initial guesses are frequently not available.
Thus, putative correspondences generated by the data association step are often contaminated with incorrect matches.
Additional effort is then required to identify which measurements are useful, as even a small number of corrupted measurements can cause an estimator to diverge~\cite{huber2004robust}.
Note that the difficulty of this requirement is in the meaning of ``useful''---in general, the classification of measurements and their correspondences as inliers or outliers is unobservable.

Conventional methods employ \emph{consensus maximization}~\cite{chin2017maximum} to reject outliers \emph{during} estimation, so that model parameters are sought that explain the largest subset of the input data.
However, this requires distinguishing between inliers and outliers via a pre-specified error threshold, which is difficult to determine in practice.
In fact, measurements which satisfy the error threshold may still be outliers in the sense that they were actually generated from a different model~\cite{carlone2014selecting}.
Alternatively, estimation in the presence of contaminated data can be performed via \emph{M-estimation}~\cite{huber1964robust}.
In this framework, the effects of outliers are discounted \emph{during} estimation by minimizing a robust loss function defined over residuals, thereby using residuals as a means of outlier detection.
This technique has been successfully used in computer vision~\cite{zhang1997parameter} and robotics~\cite{bosse2016robust} settings with a moderate number of outliers.
However, common robust losses (e.g., the convex Huber loss) have low theoretical breakdown points and often exhibit breakdown with data having even a low proportion of outliers~\cite{stewart1999robust,carlone2018convex,lajoie2019modeling}; in fact, even a \emph{single} ``bad'' outlier can lead to estimator bias~\cite{lajoie2019modeling}.

Rather than attempting to distinguish inliers from outliers, we instead aim to select the largest set of \emph{mutually consistent correspondences}, thus avoiding the unobservable nature of inlier--outlier classification and also avoiding the necessity of an initial estimation guess.
By selecting good measurements \emph{before} estimation, we focus on making the data association stage more robust, thereby increasing the quality of data before it is used for estimation.
By substantially reducing the number of incorrect correspondences in the data association module, classical robust estimation techniques become applicable again, even when the input data is heavily contaminated with outliers.
Additionally, by clipping outlier associations early in the data processing pipeline, the computational burden of processing invalid measurements during estimation is reduced.

We present a method for robust data association based on an edge-weighted graph, where edge weights encode the pairwise consistency of potential associations.
Using this graph, we formulate consistent measurement selection as a novel combinatorial optimization problem which seeks the \emph{densest} edge-weighted clique (DEWC).
Due to the computational challenges associated with identifying the DEWC in practice, we explore a continuous relaxation based on the so-called maximum spectral radius clique (MSRC).
We find that the MSRC problem has an empirically large basin of attraction with respect to the global optima of the DEWC on a variety of relevant data association problems and has connections to spectral matching~\cite{leordeanu2005spectral,olson2005single}, while being distinct in the fact that it enforces solutions to be \emph{complete} subgraphs (i.e., cliques).
In short, the use of edge weights in a density objective allows for more expressive modelling of association consistency while the clique constraint ensures that selected associations are jointly consistent, hence affording increased robustness.

To approach the non-convex MSRC problem, we introduce a convex semidefinite relaxation following Shor's relaxation~\cite{luo2010semidefinite}, enabling recovery of the globally optimal MSRC provided satisfaction of the rank constraint, notwithstanding the non-convexity of the original MSRC problem.
\rev{In practice, we find that the rank constraint is satisfied in the cases we consider in Section~\ref{sec:experiments} and that the relaxation can be solved in one second or less for problems with 100 or less putative associations.}
Finally, we present a separate, computationally-efficient MSRC solver by leveraging a homotopy-based~\cite{dunlavy2005homotopy}, graduated projected gradient ascent framework.
We find this first-order method to perform well, solving problems with 8000 associations in one second or less and frequently returning the globally-optimal solution in the problems we consider.
This algorithmic framework, called CLIPPER (Consistent LInking, Pruning, and Pairwise Error Rectification), outperforms the state-of-the-art in consistency graph formulations.

The contributions of this article can be summarized as: 
\begin{enumerate}
    \item A novel DEWC formulation for pairwise data association that provides a high degree of robustness to outlier associations as compared to existing pairwise data association techniques.
    \item The MSRC problem as a relaxation of the DEWC with an empirically large basin of attraction to the DEWC and a connection to spectral matching.
    \item A convex semidefinite relaxation of MSRC that empirically produces globally optimal solutions.
    \item A scalable, computationally efficient projected gradient ascent algorithm called CLIPPER that is shown to outperform the state-of-the-art in pairwise data association.
    \item Extensive comparisons in applications that typically do not have a reliable initial estimation guess.
\end{enumerate}
\rev{An early version of CLIPPER (contribution 4) which optimized the MSRC problem (contribution 2) was originally presented in~\cite{lusk2021clipper}.}
In addition to computational improvements and additional experimental comparisons, this article presents the theoretical foundation for our novel DEWC formulation (contribution 1), along with a specific convex semidefinite relaxation (contribution 3) that enables solving our non-convex continuous relaxation (contribution 2) to global optimality.

%% file: paper/related_work.tex
\section{Related Work}\label{sec:relatedwork}

Estimation in the presence of outliers has been widely studied in the statistics, robotics, and computer vision communities.
In the following, we discuss three leading approaches to handling outliers. %
Two major themes prevail when dealing with outliers: their effects are either alleviated or they are rejected altogether.
Often, these outlier management schemes are performed in tandem with the estimation task.
In this work, we focus on adding robustness to the data association step, \emph{rejecting} outliers \emph{before} data is processed by the estimator.

\subsection{Robust Estimation}

Maximum consensus~\cite{chin2017maximum} estimation is an NP-hard problem~\cite{chin2018robust} that aims to
estimate model parameters which are consistent with as many inliers as possible, where the inlier--outlier decision is governed by a pre-determined threshold.
RANSAC~\cite{fischler1981random} and its variants are frequently used to approximately solve this problem; however, the core hypothesize-and-verify heuristic has fundamental shortcomings.
Namely, it does not provide any certainty that the obtained result is a satisfactory approximation.
Further, such randomized methods tend to be computationally expensive in the presence of high outlier proportions because the probability of selecting an outlier-free minimal subset decreases exponentially with the number of outliers~\cite{raguram2008comparative}. %
Approximate deterministic approaches have recently been presented for the general case in~\cite{le2019deterministic} and for point cloud registration in~\cite{le2019sdrsac}.
A related approach in the context of planar SLAM is that of \citet{carlone2014selecting}, where a large set of consistent measurements is approximately sought under a linear estimation model.
\rev{Exact maximum consensus methods are commonly based on branch-and-bound~\cite{li2009consensus,yang2015goicp} or tree search~\cite{chin2015efficient}, which limits the scalability of such methods.}

Another framework for robust model fitting is M-estimation, which supplants the typical, yet highly-sensitive, least-squares loss function with a robust alternative.
The convex Huber loss function is commonly used, which has a theoretical breakdown point of 50\% and has often been known to exhibit breakdown even with less than 50\% outliers in the data~\cite{stewart1999robust,carlone2018convex}.
Non-convex loss functions exhibit increased robustness due to their redescending influence functions~\cite{holland1977robust}, such as the Geman-McClure (GM)
loss used for camera localization in~\cite{mactavish2015all} or the truncated least squares (TLS) loss used for multiple view geometry in~\cite{enqvist2012robust}, for point cloud registration in~\cite{yang2019polynomial}, and for robust pose graph optimization in~\cite{lajoie2019modeling}.
Robust losses are typically optimized locally via iteratively re-weighted least squares (IRLS)~\cite{holland1977robust}.
Thus, the use of more robust, but non-convex, loss functions can unfortunately cause the M-estimation to become (even more) sensitive to initial conditions.

An alternate strategy to solve non-convex M-estimation problems is to leverage graduated non-convexity (GNC)~\cite{blake1987visual}, a heuristic in which a convex surrogate problem is first solved, after which the problem is gradually made more non-convex---using the previous solution as the initial guess---until arriving at the original non-convex problem.
In \citet{zhou2016fast}, this strategy was employed to solve the point cloud registration problem with GM loss.
\citet{yang2020graduated} studied GNC in combination with TLS and GM for variety of spatial perception problems and found TLS to be more robust, generally up to 80\% outliers, and have applied GNC-TLS to point cloud registration~\cite{yang2020teaser}.

The success of TLS as a robust loss in spatial perception problems, but the local and heuristic nature of GNC, has led researchers to consider globally-optimal solvers and/or certifiers~\cite{carlone2022estimation}.
These algorithms typically relax the problem using Shor's relaxation~\cite{luo2010semidefinite} and solve the resulting convex semidefinite program (SDP) to compute a global solution and a numerical certificate of optimality, typically based on the rank of the SDP solution.
The certificate of optimality relies on a zero-duality gap and leads to a certifiable algorithm~\cite{bandeira2016note}.

\subsection{Correspondence Selection}

Associating elements from two sets based on inter-element distances or similarity scores is traditionally formulated as a linear assignment problem~\cite{burkard2009assignment},
which can be solved in polynomial time using, e.g., the Hungarian~\cite{kuhn1955hungarian} algorithm. %
Other pairwise association approaches include greedy techniques, such as matching feature points between two views based on nearest neighbors.
Often smart heuristics are included to detect outlier correspondences, for example, Lowe's ratio test~\cite{lowe2004sift} in feature-based image matching.
If elements have underlying structure (e.g., geometric structure like in 3D point clouds), this additional information can be used to formulate a quadratic assignment program (QAP)~\cite{lawler1963quadratic}, which is equivalent to graph matching.
Unlike linear assignment, quadratic assignment is in general NP-hard. %
Approximate solutions are thus typically sought for, e.g., using spectral graph matching~\cite{leordeanu2005spectral}.
However, QAPs do not necessarily guarantee mutual consistency of associations due to the absence of the so-called clique constraint, as explained in Section~\ref{sub:consistency}.

\subsection{Consistency Graph Approaches}\label{sub:consistency}

Consistency graph formulations seek to identify sets of pairwise consistent correspondences.
A \emph{consistency graph} is built such that vertices represent associations and edges represent the associations' pairwise consistency, much like the QAP.
Finding consistent correspondences then becomes a graph optimization problem, where the ``best'' (made more precise in Section~\ref{sec:clipper}) subgraph is sought.
These formulations assume that this ``best'' subgraph is made of inlier associations and that it stands out from ``worse'' subgraphs that are made up of noisy, bad associations.
Importantly, consistency graph approaches typically enforce that the subgraph of associations forms a clique, thus guaranteeing the mutual consistency of associations.
\citet{ambler1975versatile} presented one of the first consistency graph formulations for model-based visual part recognition using an unweighted graph.
\citet{bolles1979robust} adopted this framework allowing for weighted edges, but ultimately thresholds the edges into an unweighted graph.
More recently, \citet{bailey2000data} performed 2D LiDAR scan matching using an unweighted consistency graph formulation.
Enqvist et al.~\cite{enqvist2009optimal} developed a consistency graph method for 3D-3D and 3D-2D registration.
\citet{mangelson2018pairwise} developed the pairwise consistent measurement set maximization (PCM) algorithm for loop closure filtering, where an unweighted consistency graph encodes the consistency of loop closures in SLAM.
\rev{This approach was later extended to consider group-$k$ consistency as a generalization to pairwise consistency by Forsgren et al.~\cite{forsgren2022group,forsgren2023group}.
}
\rev{\citet{yang2020teaser} provide the TEASER++ algorithm for robust point cloud registration, where estimation in high outlier rates is feasible due to their maximum clique inlier selection step using a consistency graph~\cite{shi2020robin}.
However, inlier selection is followed by GNC-TLS for robust estimation as outliers may still be included in the maximum clique.}

An alternate approach to mining a correspondence subgraph from consistency graphs was presented in parallel by \citet{leordeanu2005spectral} and \citet{olson2005single}.
These works approach the problem from the graph matching perspective, and thus forgo the clique constraint and cannot guarantee mutually consistent associations.
Instead, they leverage edge-weighted consistency graphs to identify dense subgraphs using computationally-efficient approximations compared to Goldberg's polynomial time algorithm~\cite{goldberg1984finding}.
However, the dense subgraphs are not necessarily complete and tend to perform poorly in high-outlier settings.

In this article, we develop a consistency graph-based optimization that builds on the positive attributes of these two approaches: the clique constraint for high robustness and optimizing weighted graphs that more informatively encode continuous similarity scores.
The performance improvement caused by these two key properties is shown in Section~\ref{sec:experiments}.

%% file: paper/consistent_measurement_selection.tex
\begin{figure*}[th!]
    \centering
    \begin{subfigure}[b]{0.24\textwidth}
        \includeinkscape[pretex=\small,width=\columnwidth]{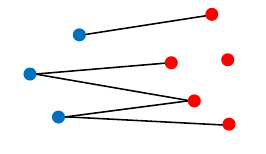}
        \caption{}
        \label{fig:example-pcds}
    \end{subfigure}
    \begin{subfigure}[b]{0.24\textwidth}
        \includeinkscape[pretex=\small,width=\columnwidth]{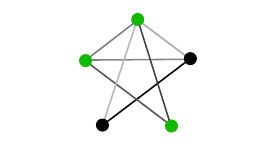}
          \caption{}
        \label{fig:example-consistency-graph}
    \end{subfigure}
    \begin{subfigure}[b]{0.24\textwidth}
        \includeinkscape[pretex=\small,width=\columnwidth]{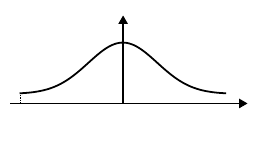}
          \caption{}
        \label{fig:example-score}
    \end{subfigure}
    \begin{subfigure}[b]{0.24\textwidth}
        \includeinkscape[pretex=\small,width=\columnwidth]{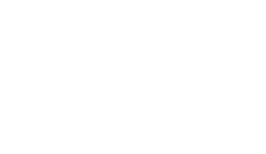}
          \caption{}
        \label{fig:example-affinity}
    \end{subfigure}
    \caption{
    Consistency graph construction example for point cloud registration.
    (a) Putative associations $u_1, \dots, u_5\in\mathcal{A}$ are given between red and blue point clouds. 
    (b) The consistency graph $\mathcal{G}$ with vertices representing the associations and edges between two vertices indicating their geometric consistency.
    In the noiseless case, any two associations $u_i,u_j$ mapping points $p_i,p_j$ to $q_i,q_j$ are consistent if $\delta=0$, where $\delta\eqdef\|p_i-p_j\| - \|q_i-q_j\|$.
    The correct associations are colored green.
    (c) Edges of the consistency graph are weighted according to the pairwise consistency score function $s(\delta)$.
    If $\delta>\epsilon$ or if two associations start/end at the same point, the association pair is deemed inconsistent.
    (d) The affinity matrix $M$ is the numerical representation of the consistency graph $\mathcal{G}$.
    }
    \vspace*{-1.5em}
    \label{fig:example}   
\end{figure*}

\section{Consistent Correspondence Selection}\label{sec:consistent-corres-sel}

Data association is hard because the classification of correspondences as inliers or outliers is unobservable.
Instead, we aim to select the group of correspondences that are most consistent with each other and the estimation model.
To do so, we first discuss the consistency graph and then present our problem formulation for consistent correspondence selection.

\subsection{Consistency Graph Construction}

Given two sets of data $\mathcal{P}, \mathcal{Q}$ and a set of putative associations $\mathcal{A}\subset\mathcal{P}\times\mathcal{Q}$, the data association problem can be viewed as a bipartite graph $\mathcal{B}=(\mathcal{P}, \mathcal{Q}, \mathcal{A})$ where $\mathcal{P}, \mathcal{Q}$ are disjoint and independent sets of vertices and $\mathcal{A}$ are the edges between $\mathcal{P}$ and $\mathcal{Q}$.
Defining $u\eqdef(p,q)\in\mathcal{A}$, a pair of associations $u_i,u_j\in\mathcal{A}$ is called \emph{consistent} if the underlying data $p_i,p_j$ can be mapped into $q_i,q_j$ using a single mapping.
However, noise prevents a perfect mapping and so a \emph{consistency score} $s:\mathcal{A}\times\mathcal{A}\to[0,1]$ is used, with $0$ being inconsistent and $1$ being fully consistent.
The scoring function typically relies on an \emph{invariant}---that is, a property that does not change under the mapping (e.g., Euclidean distance under rigid-body transformation~\cite{lusk2021clipper,shi2020robin}), though non-invariant properties have also been used (e.g., Euclidean distance under projective transformation~\cite{leordeanu2005spectral}).
By scoring each of the $\binom{|\mathcal{A}|}{2}$ association pairs, a \emph{consistency graph} $\mathcal{G}=(\mathcal{A}, \mathcal{E})$ can be formed, with associations $\mathcal{A}$ as the vertices and edges $\mathcal{E}$ weighted according to the pairwise consistency score $s$.
Note that various problem-specific constraints (i.e., element $p\in\mathcal{P}$ can only be matched to one element $q\in\mathcal{Q}$) can be enforced by removing certain edges in $\mathcal{E}$.
These constraints are especially important in the all-to-all hypothesis case, i.e., no prior information is available (e.g., no descriptor matching) and so every possible association is given in the putative association set $\mathcal{A}$.

Without loss of generality, we focus on consistency graph construction for the point cloud registration problem.
However, consistency graphs can be created for other data association and geometric estimation problems (e.g., rotation averaging~\cite{shi2020robin}, plane/line alignment~\cite{lusk2022global,lusk2022graffmatch}).
Point cloud registration is concerned with finding the rotation and translation that best align a point set to their corresponding points in another set.
Identifying correct point correspondences is the main challenge due to noise and outliers.
\textit{Outlier points} are spurious measurements, typically generated due to noisy sensing or partial observation, while \textit{outlier associations} are correspondences that incorrectly match two points from either set.
Fig.~\ref{fig:example} illustrates consistency graph construction given a set of putative associations (i.e., the lines denoted with $u_i$) between blue and red point clouds (see Fig.~\ref{fig:example-pcds}).
Inlier points are denoted by $a,b,c\in\mathcal{P}$ in the blue point cloud and their transformed counterparts by $a',b',c'\in\mathcal{Q}$ in the red point cloud.
Points $d',e'\in\mathcal{Q}$ do not correspond to any blue point and hence are considered as outliers.
Thus, the associations denoted $u_1,u_2,u_4\in\mathcal{A}$ are inliers and $u_3,u_5\in\mathcal{A}$ are outliers.

To build the consistency graph (illustrated in Fig.~\ref{fig:example-consistency-graph}), the consistency score function $s$ must be defined for point cloud registration problems.
Because rigid-body transformation (i.e., the unknown variable of the registration problem) is distance preserving, the Euclidean distance between points in one set should be identical (in the noiseless setting) to the Euclidean distance between their counterparts in the other set.
Thus, with ${\delta\eqdef\|p_i-p_j\| - \|q_i-q_j\|}$ for associations $u_i,u_j\in\mathcal{A}$ and slight abuse of notation on the domain of $s$, we define
\begin{equation}\label{eq:consistency-score}
s(\delta)\eqdef
    \begin{cases} 
        \exp(-\frac{1}{2}\frac{\delta^2}{\sigma^2}) & |\delta|\leq\epsilon \\
        0 & |\delta|>\epsilon 
    \end{cases},
\end{equation}
where the threshold $\epsilon$ is based on a bounded noise model with a noise radius of $\epsilon/2$ on point coordinates~\cite{yang2020teaser} and the parameter $\sigma$
controls how consistent an association pair with noisy underlying points is (see Fig.~\ref{fig:example-score}).
Note that consistency scores of correct association pairs are higher, which will motivate our problem formulation in Section~\ref{sec:prob-form}.

Finally, the \emph{affinity matrix} $M\in[0,1]^{m\times m}$ of the consistency graph in Fig.~\ref{fig:example-consistency-graph} is shown in Fig.~\ref{fig:example-affinity}.
The affinity matrix is an $m\times m$ symmetric matrix, where $m\eqdef|\mathcal{A}|$ is the number of putative associations.
While the off-diagonal $M_{ij}$ terms encode the pairwise consistency of associations $u_i,u_j$, the diagonal $M_{ii}$ terms encode single association consistency by measuring the similarity of points directly, for example, by comparing descriptor similarity of points $p_i, q_i$.
These terms $M_{ii}$ are simply set to $1$ when this information is unavailable.

\subsection{Problem Formulation}\label{sec:prob-form}

Given a consistency graph $\mathcal{G}$ and assuming that inlier associations $\mathcal{A}_\text{in}\subset\mathcal{A}$, our goal is to identify the subgraph $\mathcal{C}\subset\mathcal{G}$ whose vertex set is equal to $\mathcal{A}_\text{in}$.
Towards this goal, we make two observations about the properties of $\mathcal{C}$.
First, because each $(p,q)\in\mathcal{A}_\text{in}$ is such that $p$ can be mapped onto $q$ given a single mapping, then inlier associations are \emph{mutually consistent}, which forces $\mathcal{C}$ to be a clique (each vertex is connected to every other vertex). 
Second, under the assumption that noise and outliers are random and unstructured, then $\mathcal{C}$ is the ``largest'' (in some sense) clique in $\mathcal{G}$~\cite{leordeanu2005spectral}.
In particular, we propose that $\mathcal{C}$ can be identified as the \emph{densest-edge weighted clique} (DEWC) via the following optimization formulation
\begin{equation}\label{eq:dewc}
\begin{aligned}
& \underset{u\in\{0,1\}^m}{\text{maximize}} && \frac{u^\top M u}{u^\top u} \\
& \text{subject to}  & &  u_i u_j = 0 \quad \text{if}\; M_{ij}=0, \; \forall_{i,j}. \\
\end{aligned}
\end{equation}
Here, the optimization variable $u$ is a binary vector with $1$'s indicating selected associations and $0$'s otherwise.
Since $u$ is binary and the objective is to maximize, the constraint ${u_i\,u_j = 0}$ enforces the subgraph induced by $u$ to be a clique.
The objective evaluates the \emph{density} of the induced subgraph, which is defined as the total sum of edge weights divided by the number of selected vertices.
Thus, \eqref{eq:dewc} shares the objective function of the spectral matching technique~\cite{leordeanu2005spectral}, while enforcing mutual consistency via the clique constraint.

When $M$ is binary (e.g., ${s(\delta)\eqdef 1\;\text{for}\;|\delta|\leq\epsilon,\,\,0}$ otherwise and has $1$'s on the diagonal), it is straightforward to show that \eqref{eq:dewc} simplifies to the \emph{maximum clique} (MC) problem
\begin{equation}\label{eq:mc}
\begin{aligned}
& \underset{u\in\{0,1\}^m}{\text{maximize}} && \sum_{i = 1}^{m}{u_i} \\
& \text{subject to}  & &  u_i u_j = 0 \quad \text{if}\; M_{ij}=0, \; \forall_{i,j}.\\
\end{aligned}
\end{equation}
This definition of $s$ corresponds to frequently used \emph{unweighted} consistency graph frameworks~\cite{bailey2000data,enqvist2009optimal,yang2020teaser,shi2020robin}, which effectively \emph{ignore} the consistency information captured by \eqref{eq:consistency-score}.
This can be problematic in the case of \emph{competing cliques}, e.g., if the graph in Fig.~\ref{fig:example-consistency-graph} where unweighted, problem~\eqref{eq:mc} cannot disambiguate between $\{u_1,u_2,u_4\}$ (correct) or $\{u_1,u_3,u_5\}$.

To highlight the importance of density in \eqref{eq:dewc}, consider the following
affinity matrix $M$ with two solution candidates $u,u'$
\begin{equation} \label{eq:Mexample}
M =
\begin{bmatrix}
1 & 1 & 0 & 0 & 0 \\
1 & 1 & 0 & 0 & 0 \\
0 & 0 & 1 & 0.2 & 0.2 \\
0 & 0 & 0.2 & 1 & 0.2 \\
0 & 0 & 0.2 & 0.2 & 1 \\
\end{bmatrix}, 
~
u = 
\begin{bmatrix}
1 \\
1 \\
0 \\
0 \\
0 \\
\end{bmatrix},
~
u' = 
\begin{bmatrix}
0 \\
0 \\
1 \\
1 \\
1 \\
\end{bmatrix}.
\end{equation}
Both the MC objective~\eqref{eq:mc} (with non-zero $M_{ij}$ set to $1$) and the unnormalized objective of $u^\top M\, u$ return $u'$ as the optimum solution.
However, the block of $M$ corresponding to $u'$ has low pairwise consistency scores of $0.2$.
On the other hand, the density objective~\eqref{eq:dewc} takes values of $2,1.4$ for $u,u'$, respectively, leading to selection of the smaller, but more consistent subgraph. 
While this problem could have been avoided by choosing a smaller $\epsilon$ in \eqref{eq:consistency-score}, a conservative threshold may lead to rejecting correct associations (i.e., lower recall). 

\subsection{Non-Convex Continuous Relaxation}\label{sec:msrc}
A core challenge in solving \eqref{eq:dewc} is the combinatorial complexity of the problem due to its binary domain.
This makes it intractable to solve \eqref{eq:dewc} to global optimality in real time, even for small-sized problems (see Fig.~\ref{fig:scalability_timing}).
A standard approach is to relax the binary domain to the reals, which often facilitates faster optimization.
This approach is further motivated in our case due to the observation that the objective of \eqref{eq:dewc} is the Rayleigh quotient, whose maximizer over the reals is simply the principal eigenvector of $M$.
Thus, we relax the binary domain and leverage the scaling invariance of the Rayleigh quotient, yielding an optimization over the non-negative reals
\begin{equation}\label{eq:msrc}
\begin{aligned}
& \underset{v\in\mathbb{R}^m_+}{\text{maximize}} && v^\top M\,v \\
& \text{subject to} & & v_i v_j = 0 \quad \text{if}\; M_{ij}=0, \; \forall_{i,j} \\
&&&  \|v\|^2_2 \leq 1, \\
\end{aligned}
\end{equation}
which can be understood as a constrained eigenvalue problem.
In fact, due to the graph interpretation of $M$, problem \eqref{eq:msrc} seeks the \emph{maximum spectral radius clique} (MSRC) of $M$.
To recover a solution that is feasible with respect to \eqref{eq:dewc}, a \emph{rounding} step is required to project $v$ onto the binary domain.
This projection can be performed exactly by using Goldberg's polynomial-time algorithm~\cite{goldberg1984finding} to solve the \emph{densest subgraph} (DS) problem (e.g., \eqref{eq:dewc} without clique constraints) on the clique induced by the non-zero elements of $v$.

%% file: paper/clipper.tex
\section{CLIPPER}\label{sec:clipper}

Optimizing for the MSRC~\eqref{eq:msrc} would simply be the principal eigenvector if the clique constraints were omitted.
Note that because $M\in[0,1]^{m\times m}$, by the Perron-Frobenius theorem, the principal eigenvector $v^*\in[0,1]^m$, thus satisfying the non-negative real constraint~\cite{leordeanu2005spectral}.
However, the inclusion of the clique constraints makes the problem more challenging; therefore, we leverage the following penalty form
\begin{gather} \label{eq:msrc-penalty-form}
\begin{array}{ll}
\underset{v \in \mathbb{R}^m_+}{\text{maximize}} & F_d \eqdef v^\top  M_d \, v 
\\
\text{subject to} & \| v \|^2_2 \leq 1,
\end{array}
\end{gather}
where $M_d\eqdef M - d\,C$ encodes the clique constraints using the penalty parameter $d>0$ and the matrix $C\in\{0,1\}^{m\times m}$ with $C_{ij}\eqdef1$ if and only if $M_{ij}=0$.
Thus, when $C_{ij}=1$, the joint selection of $v_i,v_j$ is penalized by the amount $-2d \, v_i v_j$.
Hence, as $d$ increases the entries of solution $v$ that violate the constraints are pushed to zero.
The intuition of the penalty parameter $d$ is that of other continuation or homotopy approaches~\cite{dunlavy2005homotopy}: when $d=0$ the problem is easily solved and so the solution can be used to warm start the next optimization and so on as $d$ is incrementally increased.
At each value of $d$, we optimize \eqref{eq:msrc} using projected gradient ascent (PGA).

Given a feasible solution $v^*$ to MSRC~\eqref{eq:msrc}, let $\mathcal{G}\rvert_{v^*}\subseteq\mathcal{G}$ denote the (necessarily complete) subgraph of $\mathcal{G}$ induced by the non-zero elements of $v^*$.
As stated in Section~\ref{sec:msrc}, $v^*\in\mathbb{R}^m_+$ can be rounded to a binary $u^*\in\{0,1\}^m$ by solving the DS problem.
Instead of solving for the DS exactly, we use a fast heuristic that can immediately return a binary $u^*$ by selecting the $\hat{\omega}\eqdef\mathrm{round}( v^{*\top} M\, v^*)$ largest elements of $v^*$ as vertices of $\mathcal{G}\rvert_{u^*}\subseteq\mathcal{G}\rvert_{v^*}\subseteq\mathcal{G}$.
The justification follows from the facts that $v^{*\top} M\, v^*$ (i.e., the spectral radius of $\mathcal{G}\rvert_{v^*}$), is a tight upper bound for the graph's density~\cite{cvetkovic1980spectra} and that nonzero elements of $v^*$, (i.e., the principal eigenvector of $M\rvert_{v^*}$) represent centrality of their corresponding vertices, which is a measure of connectivity for a vertex in the graph~\cite{canright2004roles}.

These steps, i.e., 1) obtaining a solution $v^*$ of \eqref{eq:msrc} via repeated PGA and 2) estimating the densest clique $\hat{\mathcal{C}}\eqdef\mathcal{G}\rvert_{u^*}$ in $\mathcal{G}\rvert_{v^*}$ by selecting the vertices corresponding to the $\hat{\omega}$ largest elements of $v^*$, constitute the CLIPPER algorithm, which is outlined in Algorithm~\ref{alg:clipper}.
The core PGA method (Lines 6--9) utilizes backtracking line search for step size selection, \rev{with $\alpha$ initialized to $1$}, followed by a projection onto the constraint manifold.
Because solutions $v^*\in\mathbb{R}^m_+$ lie on the boundary of $\|v^*\|\leq1$, the constraint manifold can be reduced to $\mathbb{R}^m_+\cap\mathcal{S}^m$, where $\mathcal{S}^m$ is the unit sphere.
Once the inner PGA has converged for a given value of $d$, the penalty is increased and the process repeats until $\mathcal{G}\rvert_{v^*}$ satisfies the clique constraints given in $C$.

The update schedule chosen for $d$ is motivated by the desire to quickly, but carefully converge to a feasible solution.
Focusing on elements of $v$ that contribute to the violation of the clique constraints allows us to do so.
The objective gradient $\nabla_vF_d=2\,(Mv - d\,Cv)$ reveals that elements $v_i>0$ corresponding to $(Cv)_i>0$ are problematic and if increased would incur more penalty.
For each of these problematic elements $v_i$, we identify the $d$ that causes the corresponding $i$-th element of the objective gradient to vanish.
Because each $v$ is normalized after a gradient step, when other elements increase, the non-zero, non-increasing elements will diminish to zero.
We found that incrementing the penalty by the average of all such $d$'s (Lines 4, 10) produces solutions that balance convergence speed and accuracy.

\begin{algorithm}[t]
\caption{CLIPPER}
\label{alg:clipper}
\small
\begin{algorithmic}[1]
\State \textbf{Input} affinity matrix $M\in[0,1]^{m\times m}$ of consistency graph $\mathcal{G}$
\State \textbf{Output} $\hat{\mathcal{C}}\eqdef\mathcal{G}\rvert_{u^*}$, dense clique of feasible subgraph $\mathcal{G}\rvert_{v^*}\subseteq\mathcal{G}$
\State $v\gets\lambda_1(M)$ \algcomment{initialize with global solution to \eqref{eq:msrc-penalty-form} with $d=0$}
\State $d\gets \mathrm{mean}\{[Mv]/[Cv] : [Cv]>0,[v]>0\}$ \algcomment{$[\cdot]$ element-wise}
\While {clique constraints not satisfied}
    \While {$v$ not converged}
        \State $v\gets v + \alpha\nabla_v F_d$\ \algcomment{$\alpha$ via backtracking line search}
        \State $v\gets\max(v/\|v\|,0)$ \algcomment{project back onto $\mathbb{R}^m_+\cap S^m$}
    \EndWhile
    \State $d\gets d + \mathrm{mean}\{[Mv]/[Cv] : [Cv]>0,[v]>0\}$
\EndWhile
\State $\hat{\omega}\gets\mathrm{round}(v^{*\top} M\, v^*)$ \algcomment{estimate dense clique size}
\State $\hat{\mathcal{C}}\gets$ vertices corresponding to largest $\hat{\omega}$ elements of $v^*$

\end{algorithmic}
\end{algorithm}

\section{Globally Optimal CLIPPER}\label{sec:sdr}

Due to the non-convexity of \eqref{eq:msrc}, searching for the MSRC in a consistency graph $\mathcal{G}$ by optimizing \eqref{eq:msrc} may lead to local, suboptimal solutions.
\rev{These suboptimal solutions may correspond to selecting fewer consistent associations than exist or to selecting a small group of highly-consistent outliers.
Thus, the ability to identify the global solution can improve the reliability of consistency graph-based data association.
}
Toward this end, we transform \eqref{eq:msrc} into a convex semidefinite program (SDP) which has only one (global) solution by definition.
The benefit is that if a rank condition on the SDP solution is satisfied, then we have actually recovered the global optimum of \eqref{eq:msrc}.
Observing that \eqref{eq:msrc} is a quadratically-constrained quadratic program, we utilize Shor's relaxation~\cite{luo2010semidefinite}, which amounts to \emph{lifting} the decision variable to $X=vv^\top$ and discarding the implied rank-$1$ constraint, which is non-convex.
Following these steps yields %
\begin{equation}\label{eq:sdr}
\begin{aligned}
& \underset{X\in\mathbb{D}^m}{\text{maximize}} && \mathrm{Tr}(MX) \\
& \text{subject to} & & X_{ij} = 0 \quad \text{if}\; M_{ij}=0, \; \forall_{i,j} \\
&&&  \mathrm{Tr}(Z) \leq 1, \\
\end{aligned}
\end{equation}
where the doubly non-negative cone ${\mathbb{D}^m\eqdef\{X\in\mathbb{S}^m_+ | X\geq0 \}}$ and $\mathbb{S}^m_+$ is the positive semidefinite cone.
If the solution $X^*$ to \eqref{eq:sdr} has rank $1$, then $v^*$ can be recovered by the principal eigenvector of $X^*$ and $v^*$ is the global optimum of \eqref{eq:msrc}.
\rev{More details on SDP lifting approaches can be found in, e.g.,~\cite{rosen2021scalable}.}

%% file: paper/experiments.tex
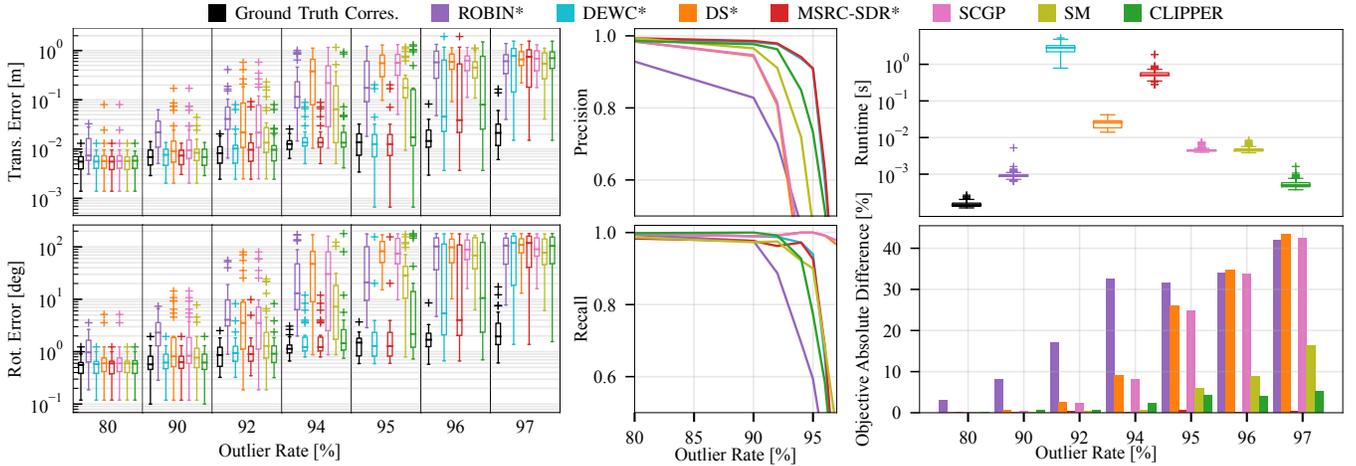
\begin{figure*}[h]
    \centering
    \begin{subfigure}[b]{1.0\textwidth}
        \centering
        \begin{tikzpicture}
            \node (origin) at (0,0) {};

            \node[right = 12mm of origin, fill={rgb,255:red,0; green,0; blue,0}, minimum size=1.5mm] (c) at (0,0) {};
            \node[right = -1mm of c] (ctxt) {\scriptsize\ \rev{Ground Truth Corres.}};
            
            \node[right = 3mm of ctxt, fill={rgb,255:red,148; green,103; blue,189}, minimum size=1.5mm] (d) {};
            \node[right = -1mm of d] (dtxt) {\scriptsize\ ROBIN*};
            
            \node[right = 3mm of dtxt, fill={rgb,255:red,23; green,190; blue,207}, minimum size=1.5mm] (e) {};
            \node[right = -1mm of e] (etxt) {\scriptsize\ DEWC*};

            \node[right = 3mm of etxt, fill={rgb,255:red,255; green,127; blue,14}, minimum size=1.5mm] (a) {};
            \node[right = -1mm of a] (atxt) {\scriptsize\ DS*};
            
            \node[right = 3mm of atxt, fill={rgb,255:red,214; green,39; blue,40}, minimum size=1.5mm] (f) {};
            \node[right = -1mm of f] (ftxt) {\scriptsize\ MSRC-SDR*};
            
            \node[right = 3mm of ftxt, fill={rgb,255:red,227; green,119; blue,194}, minimum size=1.5mm] (g) {};
            \node[right = -1mm of g] (gtxt) {\scriptsize\ SCGP};
            
            \node[right = 3mm of gtxt, fill={rgb,255:red,188; green,189; blue,34}, minimum size=1.5mm] (h) {};
            \node[right = -1mm of h] (htxt) {\scriptsize\ SM};
            
            \node[right = 3mm of htxt, fill={rgb,255:red,44; green,160; blue,44}, minimum size=1.5mm] (i) {};
            \node[right = -1mm of i] (itxt) {\scriptsize\ CLIPPER};
        \end{tikzpicture}
    \end{subfigure}
    \begin{subfigure}[b]{1.0\textwidth}
        \centering
        \includegraphics[trim = 0mm 0mm 0mm 0mm, clip, width=0.415\textwidth]{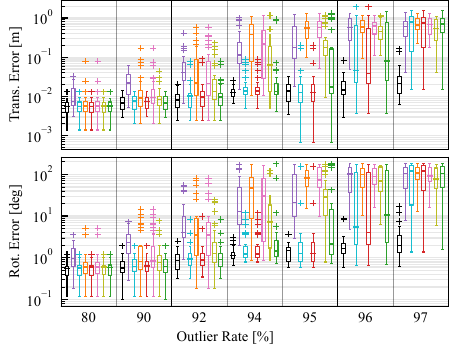}
        \includegraphics[trim = 0mm 0mm 0mm 0mm, clip, width=0.195\textwidth]{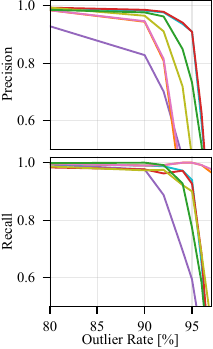}
        \includegraphics[trim = 0mm 0mm 0mm 0mm, clip, width=0.365\textwidth]{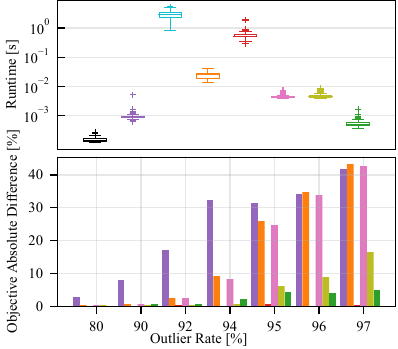}
    \end{subfigure}
    \caption{Point cloud registration in high-outlier/low-inlier regimes.
    The low registration error and high precision of DEWC* and MSRC-SDR* solutions arise by fully leveraging the weights of the consistency graph.
    CLIPPER relaxes these NP-hard problems and retains high precision and low registration error while being orders of magnitude faster.
    The low absolute difference of objective values compared to DEWC* indicates that CLIPPER frequently arrives at near-optimal solutions, while DS*, SM, SCGP fail in high outlier regimes due to the violation of the clique constraint (indicated by hash marks).
    }
    \label{fig:synthwcorres100_err}
    \vskip-0.1in
\end{figure*}

\section{Experiments}\label{sec:experiments}

We evaluate our problem formulation and algorithms in the domain of point cloud registration.
\rev{Other robotics applications (e.g., pose graph optimization, map merging, line/plane cloud matching) can also benefit from our framework, as outlined in~\cite{lusk2021clipper}.}
The DEWC~\eqref{eq:dewc} and MSRC~\eqref{eq:msrc} are solved using Gurobi~10.0.2 and the DS is solved Goldberg's polynomial-time algorithm~\cite{goldberg1984finding}, implemented in C++.
These global solutions are denoted as DEWC*, MSRC*, and DS*.
The MSRC semidefinite relaxation (MSRC-SDR)~\eqref{eq:sdr} is solved using a custom C++ parser and optimized using SCS~v3.0.0~\cite{odonoghue2016scs}.
In our experiments, we found that the the rank constraint was always satisfied for MSRC-SDR, indicating that the global optimizer of the MSRC~\eqref{eq:msrc} was found, and thus we denote its solutions as MSRC-SDR*.
Two variations of the DS are also compared against: spectral matching (SM)~\cite{leordeanu2005spectral} and single-cluster graph partitioning (SCGP)~\cite{olson2005single}, both implemented in Python.
We use the open-source, optimized C++ code of TEASER++~\cite{yang2020teaser} to test ROBIN*~\cite{shi2020robin}, which solves the MC~\eqref{eq:mc}.
TEASER++ first uses ROBIN* to reject outliers, followed by a custom TLS-GNC estimator for robust registration---we denote TLS-GNC tests without the maximum clique inlier selection stage as GNC. 
In the correspondence-based tests, we use RANSAC from Open3D~0.17 with the default desired confidence of \SI{99.9}{\percent} and denote variants limited to 10K, 100K, and 1M iterations.
Fast global registration (FGR)~\cite{zhou2016fast} is also evaluated.

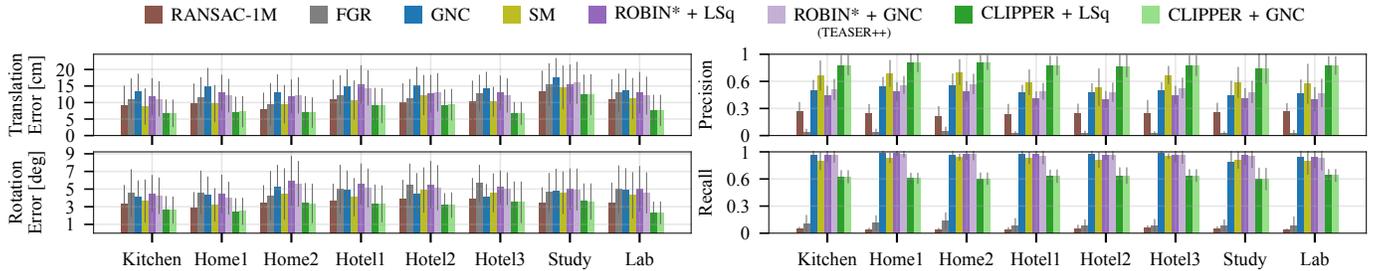
\begin{figure*}[t]
    \centering
    \begin{subfigure}[b]{1.0\textwidth}
        \centering
        \begin{tikzpicture}
            \node (origin) at (0,0) {};
            
            \node[right = 10mm of origin, fill={rgb,255:red,140; green,86; blue,75}, minimum size=1.5mm] (c) at (0,0) {};
            \node[right = -1mm of c] (ctxt) {\scriptsize\ RANSAC-1M};
            
            \node[right = 3mm of ctxt, fill={rgb,255:red,127; green,127; blue,127}, minimum size=1.5mm] (d) {};
            \node[right = -1mm of d] (dtxt) {\scriptsize\ FGR};
            
            \node[right = 3mm of dtxt, fill={rgb,255:red,31; green,119; blue,180}, minimum size=1.5mm] (e) {};
            \node[right = -1mm of e] (etxt) {\scriptsize\ GNC};

            \node[right = 3mm of etxt, fill={rgb,255:red,188; green,189; blue,34}, minimum size=1.5mm] (a) {};
            \node[right = -1mm of a] (atxt) {\scriptsize\ SM};
            
            \node[right = 3mm of atxt, fill={rgb,255:red,148; green,103; blue,189}, minimum size=1.5mm] (f) {};
            \node[right = -1mm of f] (ftxt) {\scriptsize\ ROBIN* + LSq};
            
            \node[right = 3mm of ftxt, fill={rgb,255:red,197; green,176; blue,213}, minimum size=1.5mm] (g) {};
            \node[right = -1mm of g] (gtxt) {\scriptsize\ ROBIN* + GNC};
            \node[below right = -0.75mm and 3mm of g] (gteaser) {\tiny \rev{(TEASER++)}};
            
            \node[right = 3mm of gtxt, fill={rgb,255:red,44; green,160; blue,44}, minimum size=1.5mm] (h) {};
            \node[right = -1mm of h] (htxt) {\scriptsize\ CLIPPER \rev{+ LSq}};
            
            \node[right = 3mm of htxt, fill={rgb,255:red,152; green,223; blue,138}, minimum size=1.5mm] (i) {};
            \node[right = -1mm of i] (itxt) {\scriptsize\ CLIPPER + GNC};
        \end{tikzpicture}
    \end{subfigure}
    \begin{subfigure}[b]{1\textwidth}
        \includegraphics[trim=0mm 0mm 0mm 0mm, clip, width=1\columnwidth]{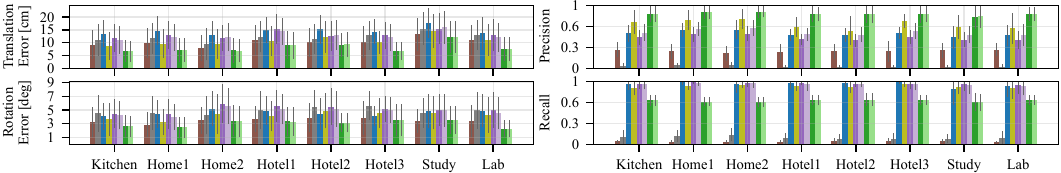}
    \end{subfigure}
    \caption{
        Average registration error, precision, and recall of successful registrations ($t_\mathrm{err}\leq\SI{30}{\centi\meter}$ and $R_\mathrm{err}\leq\SI{15}{\deg}$) from the eight 3DMatch datasets.
        Not only does CLIPPER enable the most successful registrations (cf. Table~\ref{tbl:3dmatch}), but it also produces the lowest registration error.
        This is due to its ability to achieve high precision, even in high outlier regimes.
        Because CLIPPER selects a high-precision set of correspondences, GNC does little to improve its performance.
    }
    \label{fig:3dmatch_err_and_pr}
    \vskip-0.1in
\end{figure*}

\subsection{Stanford Bunny Dataset}\label{sub:exp-bunny}

The Stanford Bunny~\cite{curless1996volumetric} model is randomly downsampled to 500 points and scaled to fit within a $[0,1]^3$ cube to obtain a source point cloud.
The target point cloud is then created by adding bounded noise to each point.
Following~\cite{yang2020teaser}, bounded noise is added by sampling ${\eta_i\sim\mathcal{N}(0,\gamma^2I)}$ and resampling if $\|\eta_i\|>\beta$.
We set $\gamma = 0.01$, with $\beta = 5.54\gamma$ chosen so that ${\mathbb{P}(\|\eta_i\|^2 > \beta^2) \leq 10^{-6}}$.
By adding noise, it is no longer clear which points correspond between the source and target point clouds.
To identify inlier correspondences, the mutual nearest neighbor bounded by $\beta$ is found for each point (when possible) in the source cloud.
False correspondences are constructed by taking the complement of the true correspondences, i.e., an all-to-all correspondence with the ground truth correspondences removed.
Putative correspondences are then simulated by combining a fraction of ground truth correspondences with false correspondences in accordance with the desired outlier rate.
Finally, a random rigid-body transform $(R,t)$ with $R\in\SO{3}$ and $t\in\Real{3}$ is applied to the target point cloud.
At each desired outlier rate, we perform 30 Monte Carlo trials.

The data association quality of algorithms is evaluated using precision and recall.
Additionally, the effect of data association on the registration task is evaluated using rotation and translation error, defined as $\|\mathrm{Log}(\hat{R}^\top R)\|$ and $\|t-\hat{t}\|$, respectively, where the estimate $(\hat{R}, \hat{t})$ is produced using Arun's (non-robust) least squares method~\cite{arun1987least}.

\textbf{Results.}
Because of the computationally intensive nature of DEWC*, we only select $m=100$ putative correspondences in the manner explained above over a range of high-outlier regimes.
Note that at \SI{97}{\percent} outlier ratio only $3$ inlier correspondences remain---this is the minimum number of points for alignment.
Registration error, precision and recall, runtime, and suboptimality results are shown in Fig.~\ref{fig:synthwcorres100_err}, \rev{where ``Ground Truth Corres.'' indicates the use of correct (but noisy) correspondences in Arun's method and is included as an indication of the best achievable estimation error.}
Precision and recall plots in Fig.~\ref{fig:synthwcorres100_err} verify that the DEWC~\eqref{eq:dewc} formulation and its relaxation MSRC~\eqref{eq:msrc} lead to robust data association, both yielding at least \SI{98}{\percent} precision until \SI{92}{\percent} outliers.
CLIPPER performs nearly as well, but with a runtime that is 3 orders of magnitude faster. 
The effect of maintaining edge weights in the consistency graph is highlighted here---in terms of precision, the next best algorithms are SM, SCGP, and DS*.
\rev{Finally, ROBIN* is the most sensitive to high outlier rates, confirming that thresholding consistency scores leads to information loss.}
As implied by the precision results, DEWC*, MSRC-SDR*, and CLIPPER achieve the best estimation error.
Finally, the suboptimality of each solution with respect to DEWC* is shown in Fig.~\ref{fig:synthwcorres100_err}, showing that (1) MSRC and CLIPPER solutions are nearly optimal and (2) that the disregarding weighted consistency information or clique constraints cause relatively higher suboptimallity.

\subsection{3DMatch Dataset}\label{sub:3dmatch}

The 3DMatch dataset~\cite{zeng20163dmatch} contains RGB-D scans of 8 indoor scenes, each scene broken into a number of fragments.
On average, each scene has approximately $200\pm135$ potential pairwise fragment registrations, as determined from the percent overlap.
The dataset provides 5000 randomly sampled keypoints and their associated 512-dimensional FPFH~\cite{rusu2009fast} features.
We randomly subselect 1000 of these keypoints, making the problem more challenging as there are fewer correct correspondences while being in high outlier regimes.
Given a pair of fragments, we generate putative correspondences by searching for 2 nearest mutual neighbors in descriptor space for each keypoint of the target fragment.
Following~\cite{zeng20163dmatch,shi2020robin,yang2020teaser}, we consider a registration successful if its error with respect to ground truth is less than \SI{15}{\deg} and \SI{30}{\centi\meter}.

\textbf{Results.}
Algorithm success rates are given in Table~\ref{tbl:3dmatch} for each dataset, where the input outlier rate and the average algorithm runtime is also shown.
In addition to using Arun's non-robust least squares method for registration, ROBIN* and CLIPPER matches are also used for GNC-based registration.
Note that ROBIN*+GNC is effectively TEASER++~\cite{yang2020teaser} and that the effect of GNC is far less significant on the CLIPPER results. 
The average precision, recall, and registration error for successful matches are shown for each algorithm in Fig.~\ref{fig:3dmatch_err_and_pr}.

\begin{table}[t]
\scriptsize
\centering
\caption{3DMatch Registration Success Rates}
\setlength{\tabcolsep}{1pt}
\ra{1.2}
\begin{tabular}{l cc cc cc cc cc cc cc cc c}
\toprule
    & \rotatebox[origin=c]{70}{Kitchen} && \rotatebox[origin=c]{70}{Home1} && \rotatebox[origin=c]{70}{Home2} && \rotatebox[origin=c]{70}{Hotel1} && \rotatebox[origin=c]{70}{Hotel2} && \rotatebox[origin=c]{70}{Hotel3} && \rotatebox[origin=c]{70}{Study} && \rotatebox[origin=c]{70}{Lab} & \begin{tabular}[x]{@{}c@{}}Average\\Runtime [ms]\end{tabular} \\
\midrule
Outlier Rate [\%] & $99.0$ && $98.7$ && $99.0$ && $99.0$ && $99.2$ && $99.3$ && $99.2$ && $99.1$ & \\
\toprule
RANSAC-10K     & $13.8$ && $16.7$ && $18.8$ && $5.8$ && $6.7$ && $5.56$ && $6.16$ && $10.4$ & $18.8  $ \\
RANSAC-100K    & $29.4$ && $46.8$ && $38.9$ && $24.8$ && $22.1$ && $38.9$ && $17.5$ && $16.9$ & $179.9  $ \\
RANSAC-1M      & $55.5$ && $60.9$ && $52.4$ && $44.2$ && $34.6$ && $46.3$ && $33.6$ && $41.6$ & $1786.6$ \\
FGR            & $50.4$ && $57.7$ && $44.2$ && $43.4$ && $31.7$ && $38.8$ && $24.7$ && $44.2$ & $12.1  $ \\
GNC            & $56.7$ && $49.4$ && $45.7$ && $47.3$ && $43.3$ && $46.3$ && $26.3$ && $50.6$ & $352.6 $ \\
SM             & $50.6$ && $60.3$ && $50.9$ && $47.3$ && $38.5$ && $40.7$ && $29.8$ && $50.6$ & $307.8 $ \\
ROBIN* + LSq   & $65.6$ && $62.2$ && $48.6$ && $57.5$ && $47.1$ && $51.9$ && $29.1$ && $54.5$ & $279.1 $ \\
ROBIN* + GNC   & $69.2$ && $64.1$ && $57.2$ && $61.5$ && $51.9$ && $55.5$ && $32.2$ && $\mathbf{55.8}$ & $284.4 $ \\
CLIPPER + LSq  & $\mathbf{77.4}$ && $74.4$ && $68.3$ && $\mathbf{69.0}$ && $\mathbf{64.4}$ && $\mathbf{70.4}$ && $45.2$ && $\mathbf{55.8}$ & $141.2 $ \\
CLIPPER + GNC  & $\mathbf{77.4}$ && $\mathbf{75.0}$ && $\mathbf{68.8}$ && $\mathbf{69.0}$ && $\mathbf{64.4}$ && $\mathbf{70.4}$ && $\mathbf{45.9}$ && $\mathbf{55.8}$ & $141.7 $ \\
\bottomrule
\end{tabular}
\label{tbl:3dmatch}
\end{table}

\begin{figure}[t]
    \centering
    \begin{subfigure}[b]{1.0\columnwidth}
        \centering
        \begin{tikzpicture}
            \node (origin) at (0,0) {};

            \node[right = 3mm of origin, fill={rgb,255:red,148; green,103; blue,189}, minimum size=1.5mm] (d) {};
            \node[right = -1mm of d] (dtxt) {\scriptsize\ ROBIN*};
            
            \node[right = 3mm of dtxt, fill={rgb,255:red,23; green,190; blue,207}, minimum size=1.5mm] (e) {};
            \node[right = -1mm of e] (etxt) {\scriptsize\ DEWC*};

            \node[right = 3mm of etxt, fill={rgb,255:red,255; green,127; blue,14}, minimum size=1.5mm] (a) {};
            \node[right = -1mm of a] (atxt) {\scriptsize\ DS*};
            
            \node[right = 3mm of atxt, fill={rgb,255:red,214; green,39; blue,40}, minimum size=1.5mm] (f) {};
            \node[right = -1mm of f] (ftxt) {\scriptsize\ MSRC-SDR*};

            \node[right = 3mm of ftxt, fill={rgb,255:red,255; green,152; blue,150}, minimum size=1.5mm] (msrc) {};
            \node[right = -1mm of msrc] (msrc-label) {\scriptsize\ MSRC*};

            \node[below right = 1mm and 15mm of origin, fill={rgb,255:red,31; green,119; blue,180}, minimum size=1.5mm] (gnc) {};
            \node[right = -1mm of gnc] (gnc-label) {\scriptsize\ GNC};
            
            \node[right = 3mm of gnc-label, fill={rgb,255:red,227; green,119; blue,194}, minimum size=1.5mm] (g) {};
            \node[right = -1mm of g] (gtxt) {\scriptsize\ SCGP};
            
            \node[right = 3mm of gtxt, fill={rgb,255:red,188; green,189; blue,34}, minimum size=1.5mm] (h) {};
            \node[right = -1mm of h] (htxt) {\scriptsize\ SM};
            
            \node[right = 3mm of htxt, fill={rgb,255:red,44; green,160; blue,44}, minimum size=1.5mm] (i) {};
            \node[right = -1mm of i] (itxt) {\scriptsize\ CLIPPER};
        \end{tikzpicture}
    \end{subfigure}
    \begin{subfigure}[b]{1\columnwidth}
        \includegraphics[trim=0mm 0mm 0mm 0mm, clip, width=1\columnwidth]{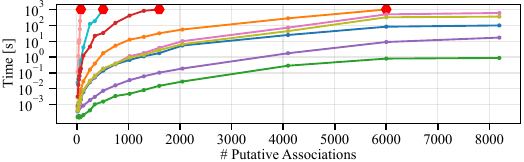}
    \end{subfigure}
    \caption{
        Scalability results.
        MSRC*, DEWC*, MSRC-SDR, and DS* had to be early stopped, indicated via the symbol \tikzhexagon.
    }
    \label{fig:scalability_timing}
\end{figure}
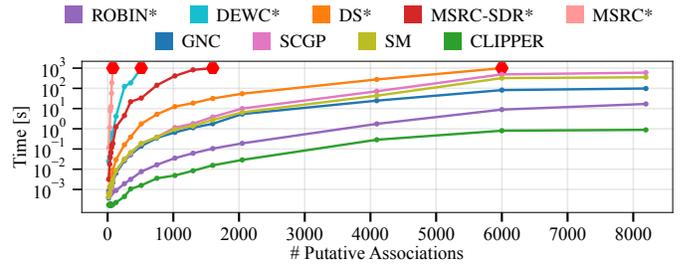

\subsection{Scalability Evaluation}\label{sub:exp-scalability}
Fig.~\ref{fig:scalability_timing} shows how the runtime of each algorithm scales with the number of putative associations, which were generated in the same way as described in Section~\ref{sub:exp-bunny} for an outlier ratio of \SI{80}{\percent} such that each algorithm had an average of \SI{90}{\percent} precision.
Runtime results are generated on a single CPU core and are averaged over 10 trials, except for MSRC*, DEWC*, and MSRC-SDR*, which are averaged over two trials.
These results indicate that CLIPPER can process 8000 associations in one second or less, achieving orders of magnitude faster performance compared to the state of the art.

%% file: paper/conclusion.tex
\section{Conclusion}\label{sec:conclusion}

\rev{
Graph-theoretic data association is a key enabler of robust perception. %
Using a \emph{weighted} consistency graph, we show that our DEWC formulation yields extreme robustness even when few inliers are present.
In particular, correspondences selected by the DEWC lead to the lowest point cloud registration error compared to commonly used formulations.
For improved efficiency, we develop the CLIPPER algorithm to optimize a DEWC approximation, which frequently produces solutions that are near-optimal with respect to the DEWC.
A convex SDP was also introduced, allowing global solutions of the DEWC to be found provided a rank constraint holds.
Future work will explore the effect of noise on the tightness of this relaxation.
}
Note that while we evaluated CLIPPER in the context of point cloud registration, our method can be applied so long as a feature between data samples can be found that is invariant to the transformation affecting the two data sets.

%% file: main.bbl
\newcommand{\noopsort}[1]{} \newcommand{\printfirst}[2]{#1}
  \newcommand{\singleletter}[1]{#1} \newcommand{\switchargs}[2]{#2#1}
\begin{thebibliography}{56}
\providecommand{\natexlab}[1]{#1}
\providecommand{\url}[1]{#1}
\csname url@samestyle\endcsname
\providecommand{\newblock}{\relax}
\providecommand{\bibinfo}[2]{#2}
\providecommand{\BIBentrySTDinterwordspacing}{\spaceskip=0pt\relax}
\providecommand{\BIBentryALTinterwordstretchfactor}{4}
\providecommand{\BIBentryALTinterwordspacing}{\spaceskip=\fontdimen2\font plus
\BIBentryALTinterwordstretchfactor\fontdimen3\font minus
  \fontdimen4\font\relax}
\providecommand{\BIBforeignlanguage}[2]{{%
\expandafter\ifx\csname l@#1\endcsname\relax
\typeout{** WARNING: IEEEtranN.bst: No hyphenation pattern has been}%
\typeout{** loaded for the language `#1'. Using the pattern for}%
\typeout{** the default language instead.}%
\else
\language=\csname l@#1\endcsname
\fi
#2}}
\providecommand{\BIBdecl}{\relax}
\BIBdecl

\bibitem[Besl and McKay(1992)]{besl1992method}
P.~J. Besl and N.~D. McKay, ``Method for registration of 3-d shapes,'' in
  \emph{Sensor fusion IV: control paradigms and data structures}, vol.
  1611.\hskip 1em plus 0.5em minus 0.4em\relax International Society for Optics
  and Photonics, 1992, pp. 586--606.

\bibitem[Huber(2004)]{huber2004robust}
P.~J. Huber, \emph{Robust statistics}.\hskip 1em plus 0.5em minus 0.4em\relax
  John Wiley \& Sons, 2004, vol. 523.

\bibitem[Chin and Suter(2017)]{chin2017maximum}
T.-J. Chin and D.~Suter, ``The maximum consensus problem: recent algorithmic
  advances,'' \emph{Synthesis Lectures on Computer Vision}, vol.~7, no.~2, pp.
  1--194, 2017.

\bibitem[Carlone et~al.(2014)Carlone, Censi, and
  Dellaert]{carlone2014selecting}
L.~Carlone, A.~Censi, and F.~Dellaert, ``Selecting good measurements via l1
  relaxation: A convex approach for robust estimation over graphs,'' in
  \emph{IEEE/RSJ IROS}, 2014, pp. 2667--2674.

\bibitem[Huber(1964)]{huber1964robust}
P.~J. Huber, ``{Robust Estimation of a Location Parameter}.''\hskip 1em plus
  0.5em minus 0.4em\relax Institute of Mathematical Statistics, 1964, vol.~35,
  no.~1, pp. 73--101.

\bibitem[Zhang(1997)]{zhang1997parameter}
Z.~Zhang, ``Parameter estimation techniques: A tutorial with application to
  conic fitting,'' \emph{IMAVIS}, vol.~15, no.~1, pp. 59--76, 1997.

\bibitem[Bosse et~al.(2016)Bosse, Agamennoni, and
  Gilitschenski]{bosse2016robust}
M.~Bosse, G.~Agamennoni, and I.~Gilitschenski, ``Robust estimation and
  applications in robotics,'' \emph{Found. and Trends in Robotics}, vol.~4, pp.
  225--269, 2016.

\bibitem[Stewart(1999)]{stewart1999robust}
C.~V. Stewart, ``Robust parameter estimation in computer vision,'' \emph{SIAM
  review}, vol.~41, no.~3, pp. 513--537, 1999.

\bibitem[Carlone and Calafiore(2018)]{carlone2018convex}
L.~Carlone and G.~C. Calafiore, ``Convex relaxations for pose graph
  optimization with outliers,'' \emph{IEEE RA-L}, pp. 1160--1167, 2018.

\bibitem[Lajoie et~al.(2019)Lajoie, Hu, Beltrame, and
  Carlone]{lajoie2019modeling}
P.-Y. Lajoie, S.~Hu, G.~Beltrame, and L.~Carlone, ``Modeling perceptual
  aliasing in slam via discrete--continuous graphical models,'' \emph{IEEE
  Robotics and Automation Letters}, vol.~4, no.~2, pp. 1232--1239, 2019.

\bibitem[Leordeanu and Hebert(2005)]{leordeanu2005spectral}
M.~Leordeanu and M.~Hebert, ``A spectral technique for correspondence problems
  using pairwise constraints,'' in \emph{IEEE International Conference on
  Computer Vision}, vol.~2, 2005, pp. 1482--1489.

\bibitem[Olson et~al.(2005)Olson, Walter, Teller, and Leonard]{olson2005single}
E.~Olson, M.~R. Walter, S.~J. Teller, and J.~J. Leonard, ``Single-cluster
  spectral graph partitioning for robotics applications.'' in \emph{Robotics:
  Science and Systems}, 2005, pp. 265--272.

\bibitem[Luo et~al.(2010)Luo, Ma, So, Ye, and Zhang]{luo2010semidefinite}
Z.-Q. Luo, W.-K. Ma, A.~M.-C. So, Y.~Ye, and S.~Zhang, ``Semidefinite
  relaxation of quadratic optimization problems,'' \emph{IEEE Signal Processing
  Magazine}, vol.~27, no.~3, pp. 20--34, 2010.

\bibitem[Dunlavy and O'Leary(2005)]{dunlavy2005homotopy}
D.~M. Dunlavy and D.~P. O'Leary, ``Homotopy optimization methods for global
  optimization.'' Sandia National Laboratories, Tech. Rep., 2005.

\bibitem[Lusk et~al.(2021)Lusk, Fathian, and How]{lusk2021clipper}
P.~C. Lusk, K.~Fathian, and J.~P. How, ``{CLIPPER: A Graph-Theoretic Framework
  for Robust Data Association},'' in \emph{IEEE ICRA}, 2021.

\bibitem[Chin et~al.(2018)Chin, Cai, and Neumann]{chin2018robust}
T.-J. Chin, Z.~Cai, and F.~Neumann, ``Robust fitting in computer vision: Easy
  or hard?'' in \emph{ECCV}, 2018, pp. 701--716.

\bibitem[Fischler and Bolles(1981)]{fischler1981random}
M.~A. Fischler and R.~C. Bolles, ``Random sample consensus: a paradigm for
  model fitting with applications to image analysis and automated
  cartography,'' \emph{Comm. of the ACM}, vol.~24, no.~6, pp. 381--395, 1981.

\bibitem[Raguram et~al.(2008)Raguram, Frahm, and
  Pollefeys]{raguram2008comparative}
R.~Raguram, J.-M. Frahm, and M.~Pollefeys, ``A comparative analysis of {RANSAC}
  techniques leading to adaptive real-time random sample consensus,'' in
  \emph{ECCV}.\hskip 1em plus 0.5em minus 0.4em\relax Springer, 2008, pp.
  500--513.

\bibitem[Le et~al.(2019{\natexlab{a}})Le, Chin, Eriksson, Do, and
  Suter]{le2019deterministic}
H.~M. Le, T.-J. Chin, A.~Eriksson, T.-T. Do, and D.~Suter, ``Deterministic
  approximate methods for maximum consensus robust fitting,'' \emph{IEEE
  T-PAMI}, 2019.

\bibitem[Le et~al.(2019{\natexlab{b}})Le, Do, Hoang, and Cheung]{le2019sdrsac}
H.~M. Le, T.-T. Do, T.~Hoang, and N.-M. Cheung, ``{SDRSAC}: Semidefinite-based
  randomized approach for robust point cloud registration without
  correspondences,'' in \emph{IEEE/CVF CVPR}, 2019, pp. 124--133.

\bibitem[Li(2009)]{li2009consensus}
H.~Li, ``Consensus set maximization with guaranteed global optimality for
  robust geometry estimation,'' in \emph{IEEE ICCV}, 2009, pp. 1074--1080.

\bibitem[Yang et~al.(2015)Yang, Li, Campbell, and Jia]{yang2015goicp}
J.~Yang, H.~Li, D.~Campbell, and Y.~Jia, ``{Go-ICP}: A globally optimal
  solution to 3d icp point-set registration,'' \emph{IEEE T-PAMI}, vol.~38,
  no.~11, pp. 2241--2254, 2015.

\bibitem[Chin et~al.(2015)Chin, Purkait, Eriksson, and
  Suter]{chin2015efficient}
T.-J. Chin, P.~Purkait, A.~Eriksson, and D.~Suter, ``Efficient globally optimal
  consensus maximisation with tree search,'' in \emph{IEEE/CVF CVPR}, 2015, pp.
  2413--2421.

\bibitem[Holland and Welsch(1977)]{holland1977robust}
P.~W. Holland and R.~E. Welsch, ``Robust regression using iteratively
  reweighted least-squares,'' \emph{Communications in Statistics-theory and
  Methods}, vol.~6, no.~9, pp. 813--827, 1977.

\bibitem[MacTavish and Barfoot(2015)]{mactavish2015all}
K.~MacTavish and T.~D. Barfoot, ``At all costs: A comparison of robust cost
  functions for camera correspondence outliers,'' in \emph{IEEE CRV}, 2015.

\bibitem[Enqvist et~al.(2012)Enqvist, Ask, Kahl, and
  Astr{\"o}m]{enqvist2012robust}
O.~Enqvist, E.~Ask, F.~Kahl, and K.~Astr{\"o}m, ``Robust fitting for multiple
  view geometry,'' in \emph{ECCV}.\hskip 1em plus 0.5em minus 0.4em\relax
  Springer, 2012, pp. 738--751.

\bibitem[Yang and Carlone(2019)]{yang2019polynomial}
H.~Yang and L.~Carlone, ``A polynomial-time solution for robust registration
  with extreme outlier rates,'' in \emph{RSS}, 2019.

\bibitem[Blake and Zisserman(1987)]{blake1987visual}
A.~Blake and A.~Zisserman, \emph{Visual reconstruction}.\hskip 1em plus 0.5em
  minus 0.4em\relax MIT press, 1987.

\bibitem[Zhou et~al.(2016)Zhou, Park, and Koltun]{zhou2016fast}
Q.-Y. Zhou, J.~Park, and V.~Koltun, ``Fast global registration,'' in
  \emph{ECCV}.\hskip 1em plus 0.5em minus 0.4em\relax Springer, 2016, pp.
  766--782.

\bibitem[Yang et~al.(2020{\natexlab{a}})Yang, Antonante, Tzoumas, and
  Carlone]{yang2020graduated}
H.~Yang, P.~Antonante, V.~Tzoumas, and L.~Carlone, ``Graduated non-convexity
  for robust spatial perception: From non-minimal solvers to global outlier
  rejection,'' \emph{IEEE RA-L}, vol.~5, no.~2, pp. 1127--1134, 2020.

\bibitem[Yang et~al.(2020{\natexlab{b}})Yang, Shi, and Carlone]{yang2020teaser}
H.~Yang, J.~Shi, and L.~Carlone, ``Teaser: Fast and certifiable point cloud
  registration,'' \emph{IEEE Transactions on Robotics}, 2020.

\bibitem[Carlone(2023)]{carlone2022estimation}
L.~Carlone, ``Estimation contracts for outlier-robust geometric perception,''
  \emph{Foundations and Trends{\textregistered} in Robotics}, vol.~11, no. 2-3,
  pp. 90--224, 2023.

\bibitem[Bandeira(2016)]{bandeira2016note}
A.~S. Bandeira, ``A note on probably certifiably correct algorithms,''
  \emph{Comptes Rendus Mathematique}, vol. 354, no.~3, pp. 329--333, 2016.

\bibitem[Burkard et~al.(2009)Burkard, Dell’Amico, and
  Martello]{burkard2009assignment}
R.~Burkard, M.~Dell’Amico, and S.~Martello, ``Assignment problems,''
  \emph{SIAM}, 2009.

\bibitem[Kuhn(1955)]{kuhn1955hungarian}
H.~W. Kuhn, ``The {Hungarian} method for the assignment problem,'' \emph{Naval
  Research Logistics Quarterly}, vol.~2, no. 1‐2, pp. 83--97, 1955.

\bibitem[Lowe(2004)]{lowe2004sift}
D.~G. Lowe, ``Distinctive image features from scale-invariant keypoints,''
  \emph{IJCV}, vol.~60, no.~2, pp. 91--110, 2004.

\bibitem[Lawler(1963)]{lawler1963quadratic}
E.~L. Lawler, ``The quadratic assignment problem,'' \emph{Management science},
  vol.~9, no.~4, pp. 586--599, 1963.

\bibitem[Ambler et~al.(1975)Ambler, Barrow, Brown, Burstall, and
  Popplestone]{ambler1975versatile}
A.~P. Ambler, H.~G. Barrow, C.~M. Brown, R.~M. Burstall, and R.~J. Popplestone,
  ``A versatile system for computer-controlled assembly,'' \emph{Artificial
  Intelligence}, vol.~6, no.~2, pp. 129--156, 1975.

\bibitem[Bolles(1979)]{bolles1979robust}
R.~C. Bolles, ``Robust feature matching through maximal cliques,'' in
  \emph{Imaging Applications for Automated Industrial Inspection and Assembly},
  vol. 182.\hskip 1em plus 0.5em minus 0.4em\relax SPIE, 1979, pp. 140--149.

\bibitem[Bailey et~al.(2000)Bailey, Nebot, Rosenblatt, and
  Durrant-Whyte]{bailey2000data}
T.~Bailey, E.~M. Nebot, J.~Rosenblatt, and H.~F. Durrant-Whyte, ``Data
  association for mobile robot navigation: A graph theoretic approach,'' in
  \emph{IEEE ICRA}, vol.~3, 2000, pp. 2512--2517.

\bibitem[Enqvist et~al.(2009)Enqvist, Josephson, and Kahl]{enqvist2009optimal}
O.~Enqvist, K.~Josephson, and F.~Kahl, ``Optimal correspondences from pairwise
  constraints,'' in \emph{IEEE/CVF ICCV}, 2009, pp. 1295--1302.

\bibitem[Mangelson et~al.(2018)Mangelson, Dominic, Eustice, and
  Vasudevan]{mangelson2018pairwise}
J.~G. Mangelson, D.~Dominic, R.~M. Eustice, and R.~Vasudevan, ``Pairwise
  consistent measurement set maximization for robust multi-robot map merging,''
  in \emph{IEEE ICRA}, 2018, pp. 2916--2923.

\bibitem[Forsgren et~al.(2022)Forsgren, Vasudevan, Kaess, McLain, and
  Mangelson]{forsgren2022group}
B.~Forsgren, R.~Vasudevan, M.~Kaess, T.~W. McLain, and J.~G. Mangelson,
  ``Group-$ k $ consistent measurement set maximization for robust outlier
  detection,'' in \emph{IEEE/RSJ IROS}, 2022, pp. 4849--4856.

\bibitem[Forsgren et~al.(2023)Forsgren, Vasudevan, Kaess, McLain, and
  Mangelson]{forsgren2023group}
------, ``Group-$ k $ consistent measurement set maximization via maximum
  clique over k-uniform hypergraphs for robust multi-robot map merging,''
  \emph{arXiv preprint arXiv:2308.02674}, 2023.

\bibitem[Shi et~al.(2021)Shi, Yang, and Carlone]{shi2020robin}
J.~Shi, H.~Yang, and L.~Carlone, ``Robin: a graph-theoretic approach to reject
  outliers in robust estimation using invariants,'' in \emph{IEEE ICRA}, 2021,
  pp. 13\,820--13\,827.

\bibitem[Goldberg(1984)]{goldberg1984finding}
A.~V. Goldberg, \emph{Finding a maximum density subgraph}.\hskip 1em plus 0.5em
  minus 0.4em\relax University of California Berkeley, 1984.

\bibitem[Lusk and How(2022)]{lusk2022global}
P.~C. Lusk and J.~P. How, ``{Global data association for SLAM with 3D
  Grassmannian manifold objects},'' in \emph{IEEE/RSJ IROS}, 2022, pp.
  4463--4470.

\bibitem[Lusk et~al.(2022)Lusk, Parikh, and How]{lusk2022graffmatch}
P.~C. Lusk, D.~Parikh, and J.~P. How, ``{GraffMatch: Global Matching of 3D
  Lines and Planes for Wide Baseline LiDAR Registration},'' \emph{IEEE RA-L},
  vol.~8, no.~2, pp. 632--639, 2022.

\bibitem[Cvetkovi\'{c} et~al.(1980)Cvetkovi\'{c}, Doob, and
  Sachs]{cvetkovic1980spectra}
D.~M. Cvetkovi\'{c}, M.~Doob, and H.~Sachs, \emph{Spectra of Graphs: Theory and
  Application}.\hskip 1em plus 0.5em minus 0.4em\relax New York: Academic
  Press, 1980.

\bibitem[Canright and Eng{\o}-Monsen(2004)]{canright2004roles}
G.~Canright and K.~Eng{\o}-Monsen, ``Roles in networks,'' \emph{Science of
  Computer Programming}, vol.~53, no.~2, pp. 195--214, 2004.

\bibitem[Rosen(2021)]{rosen2021scalable}
D.~M. Rosen, ``Scalable low-rank semidefinite programming for certifiably
  correct machine perception,'' in \emph{Proc. of the 14th Workshop on the
  Algo. Foundations of Robotics}.\hskip 1em plus 0.5em minus 0.4em\relax
  Springer, 2021, pp. 551--566.

\bibitem[O'Donoghue et~al.(2016)O'Donoghue, Chu, Parikh, and
  Boyd]{odonoghue2016scs}
B.~O'Donoghue, E.~Chu, N.~Parikh, and S.~Boyd, ``Conic optimization via
  operator splitting and homogeneous self-dual embedding,'' \emph{J. Optim.
  Theory Appl.}, vol. 169, no.~3, pp. 1042--1068, June 2016.

\bibitem[Curless and Levoy(1996)]{curless1996volumetric}
B.~Curless and M.~Levoy, ``A volumetric method for building complex models from
  range images,'' in \emph{Proc. ACM Comput. Graph. Interact. Tech.}, 1996, pp.
  303--312.

\bibitem[Arun et~al.(1987)Arun, Huang, and Blostein]{arun1987least}
K.~S. Arun, T.~S. Huang, and S.~D. Blostein, ``{Least-squares fitting of two
  3-D point sets},'' \emph{IEEE T-PAMI}, no.~5, pp. 698--700, 1987.

\bibitem[Zeng et~al.(2017)Zeng, Song, Nie{\ss}ner, Fisher, Xiao, and
  Funkhouser]{zeng20163dmatch}
A.~Zeng, S.~Song, M.~Nie{\ss}ner, M.~Fisher, J.~Xiao, and T.~Funkhouser,
  ``{3DMatch: Learning Local Geometric Descriptors from RGB-D
  Reconstructions},'' in \emph{CVPR}, 2017.

\bibitem[Rusu et~al.(2009)Rusu, Blodow, and Beetz]{rusu2009fast}
R.~B. Rusu, N.~Blodow, and M.~Beetz, ``{Fast point feature histograms (FPFH)
  for 3D registration},'' in \emph{IEEE ICRA}, 2009, pp. 3212--3217.

\end{thebibliography}
